\documentclass[runningheads]{llncs}
\usepackage[T1]{fontenc}
\usepackage{graphicx}
\usepackage{multirow}
\usepackage{amsmath}
\usepackage{booktabs,subcaption,amsfonts,dcolumn}
\newcolumntype{d}[1]{D..{#1}}
\usepackage{xcolor,colortbl}
\usepackage[misc,geometry]{ifsym}
\definecolor{Gray}{gray}{0.85}

\newcolumntype{a}{>{\columncolor{Gray}}c}
\begin{document}
\title{SeATrans: Learning Segmentation-Assisted diagnosis model via Transformer}
%
\author{Junde Wu\inst{1}, 
Huihui Fang\inst{1}, 
Fangxin Shang\inst{1}, Dalu Yang\inst{1}, Zhaowei Wang\inst{1}, Jing Gao\inst{2}, Yehui Yang\inst{1}, Yanwu Xu\inst{1}\textsuperscript{ (\Letter)}}
\authorrunning{J. Wu et al.}

\institute{Intelligent Healthcare Unit, Baidu Inc.\\
\email{ywxu@ieee.org}
\\
\and 
Purdue University\\}
\maketitle              
\begin{abstract}
Clinically, the accurate annotation of lesions/tissues can significantly facilitate the disease diagnosis. For example, the segmentation of optic disc/cup (OD/OC) on fundus image would facilitate the glaucoma diagnosis, the segmentation of skin lesions on dermoscopic images is helpful to the melanoma diagnosis, etc. With the advancement of deep learning techniques, a wide range of methods proved the lesions/tissues segmentation can also facilitate the automated disease diagnosis models. However, existing methods are limited in the sense that they can only capture static regional correlations in the images. Inspired by the global and dynamic nature of Vision Transformer, in this paper, we propose Segmentation-Assisted diagnosis Transformer (SeATrans) to transfer the segmentation knowledge to the disease diagnosis network. Specifically, we first propose an asymmetric multi-scale interaction strategy to correlate each single low-level diagnosis feature with multi-scale segmentation features. Then, an effective strategy called SeA-block is adopted to vitalize diagnosis feature via correlated segmentation features. To model the segmentation-diagnosis interaction, SeA-block first embeds the diagnosis feature based on the segmentation information via the encoder, and then transfers the embedding back to the diagnosis feature space by a decoder. Experimental results demonstrate that SeATrans surpasses a wide range of state-of-the-art (SOTA) segmentation-assisted diagnosis methods on several disease diagnosis tasks.  
\end{abstract}
\section{Introduction}\label{sec:introduction}

Clinically, the disease diagnosis is usually conducted based on critical biomarkers derived from an analysis of the images. For example, on fundus images, the vertical Cup-to-Disc Ratio (vCDR) parameter computed from the optic cup/disc  (OD/OC) masks is one of the most important clinical parameters for the glaucoma diagnosis \cite{garway1998vertical}. In melanoma diagnosis, an unusual shape of the skin lesions is a major biomarker indicating melanoma \cite{gachon2005first}. In order to derive these important biomarkers, an essential step is to identify lesions or tissues in an image and segment these areas of interest from the rest of the image \cite{ji2021learning,fu2018joint,yuan2017automatic}. 

Motivated by this observation, methods have been proposed to utilize segmentation information to facilitate the automated disease diagnosis  \cite{fu2018disc,zhou2019collaborative,wu2020leveraging,li2019attention,bajwa2019two,wu2022gamma,Yang2021RobustCL}. The common practices include region of interest (ROI) extraction \cite{fu2018disc,bajwa2019two}, input concatenation, channel attention  \cite{li2019attention,zhou2019collaborative}, and transfer learning \cite{wu2020leveraging}. These methods have two main limitations. First, the methods proposed for specific medical tasks are not general enough. They are often inapplicable or have unsatisfactory performance on other medical tasks. Second, most methods simply assume that the segmentation and diagnosis features are regional correlated, which is an invalid assumption in most cases. Traditional techniques they applied, like convolution layers and channel attentions are difficult to model this non-regional feature interaction, since these tools are largely local-focused.
With the rise of vision transformer  \cite{dosovitskiy2020image}, such a research gap can be possibly addressed by its global and dynamic nature \cite{naseer2021intriguing}.  

In this paper, we propose a novel transformer-based model to better capture the interaction of segmentation and diagnosis features. In order to address the scale-level discrepancy between segmentation and diagnosis features, we propose asymmetric multi-scale interaction to correlate multi-scale segmentation features with each single low-level diagnosis feature. A one-to-one coarse interaction and a one-against-rest fine-grain interaction are fused to produce the final feature. An effective approach, called SeA-block, is proposed to model the segmentation-diagnosis interaction, which is constructed by an encoder-decoder pair. The encoder first embeds the diagnosis feature through the calculated segmentation affinity map. Then a decoder maps the embeddings back to the diagnosis feature space through the calculated diagnosis affinity map. Through SeA-block, diagnosis features can be vitalized by the correlated segmentation information.

In brief, we have made three major contributions. First, we propose a general segmentation-assisted diagnosis model, named SeATrans, for integrating segmentation and diagnosis based on medical images. Thanks to the global and dynamic nature of transformer mechanism, SeATrans can achieve superior and robust performance comparing with state-of-the-art counterparts. Second, we propose asymmetric multi-scale interaction to correlate each low-level diagnosis feature with multi-scale segmentation features. In this way, diagnosis feature can be vitalized by both coarse and fine-grain segmentation information. Last but not the least, we propose a new strategy, i.e., SeA-block, for the segmentation-diagnosis interaction. A transformer-based encoder-decoder architecture is constructed to learn across the segmentation and diagnosis feature space. The experimental results show SeATrans outperforms previous best method by at least a 2 \% AUC over three different disease diagnosis tasks. Meanwhile, it shows competitive robustness to the domain shift of segmentation model.

\section{Methodology}
\begin{figure}[t]
\centering
\includegraphics[width=\linewidth]{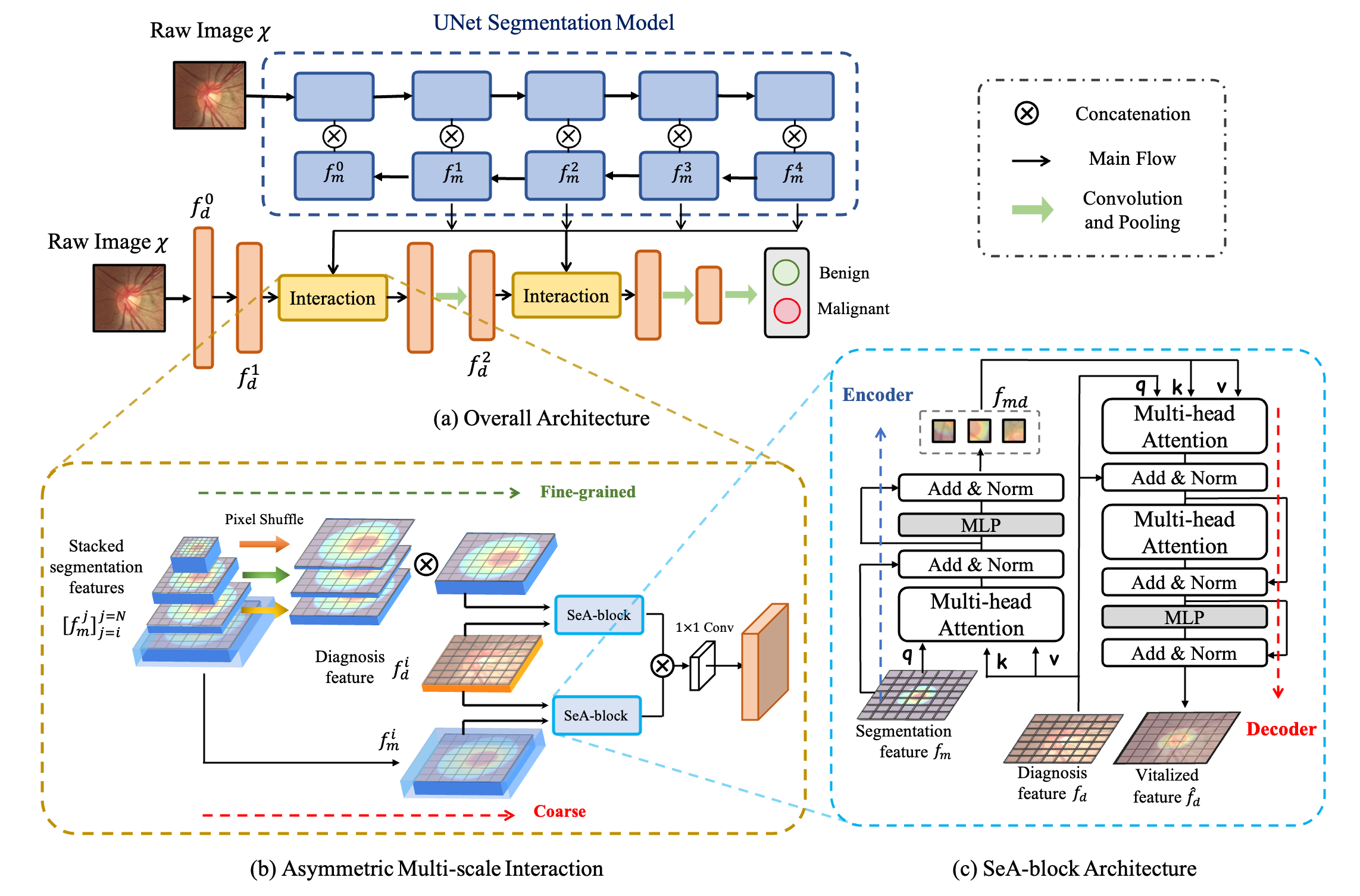}
\caption{An illustration of SeATrans framework, which starts from (a) an overview of the processing pipeline, and continues with zoomed-in diagrams of individual modules, including (b) the Asymmetric Multi-scale Interaction and (c) the SeA-block.}
\label{fig:overall}

\end{figure}

In this paper, we propose a general segmentation-assisted diagnosis framework. Given a raw image $x$ and its lesions/tissues segmentation features $f_{m}$ extracted from a segmentation network (joint-trained or pre-trained), our goal is to predict the disease $y$ (0 for benign, 1 for malignant) of the image. Our basic idea is to integrate the segmentation information into diagnosis model on the feature level. The interaction module and diagnosis model are jointly optimized to predict the correct diagnosis. An illustration of the overflow is shown in Fig. \ref{fig:overall} (a). Raw fundus image $x$ is first sent into a UNet to obtain the deep segmentation embedding. The segmentation features in the UNet decoder are used to interact with the diagnosis features of a disease diagnosis network. In the diagnosis model, convolution layer and SeA-block based Interaction alternatively abstracts and vitalizes the features. The final disease probability is supervised by the binary disease label through binary cross-entropy (BCE) loss function.

\subsection{Asymmetric Multi-scale Interaction}

Note that the diagnosis network abstracts the low-level structure features to the deep semantic features, while the segmentation model abstracts multi-scale structural features. In order to align the diagnosis and segmentation features, we correlate multi-scale segmentation features to each single low-level diagnosis feature. As shown in Fig. \ref{fig:overall} (b), stacked multi-scale segmentation features are collected for a single low-level diagnosis feature. The segmentation feature with the largest scale will first interact with the target diagnosis feature. As large-scale feature contains more specific but artifact structure information \cite{wojna2017devil}, this one-to-one interaction will produce a coarse vitalized diagnosis feature. Other segmentation features with smaller scales are fused together for the interaction with the diagnosis feature. Since these features contain more fine-grained and abstract features, this interaction will produce a fine-grained vitalized diagnosis feature. The coarse and fine-grained features are fused by $1 \times 1$ convolution layer to produce the final result.

In practice, the second and third layers of the diagnosis model will interact with the multi-scale features in UNet decoder. Consider the deep segmentation feature and diagnosis feature are $f_{m}$ and $f_{d}$. To instill segmentation information into $f^{i}_{d}\in \mathbb{R}^{\frac{H}{r} \times \frac{W}{r} \times C}$ ($i$ is the index of layer, $H$,$W$,$r$,$C$ are height, width, down-sample rate and channel number respectively), stacked multi-scale segmentation features $[f^{j}_{m}]_{j = i}^{j = N}$ ($N$ is the number of layers) are collected for the interaction. First, $f^{i}_{m}\in \mathbb{R}^{\frac{H}{r} \times \frac{W}{r} \times C}$ will interact with $f^{i}_{d}$ by SeA-block for coarse vitalization. Then the subsequent segmentation features $[f^{j}_{m}]_{j = i+1}^{j = N}$ will be rearranged by pixel shuffle \cite{shi2016real} to the scale of $\frac{H}{r} \times \frac{W}{r}$ and concatenated together. Then it will interact with diagnosis feature $f^{i}_{d}$ for the fine-grained interaction. 
The fine-grained feature and coarse feature are integrated by $1 \times 1$ convolution kernel to obtain the final vitalized diagnosis feature with shape $\frac{H}{r} \times \frac{W}{r} \times C$. Then a residual convolution block \cite{he2016deep} with pooling layer is connected to abstract the next feature $f_{d}^{i+1} \in \mathbb{R}^{\frac{H}{2r} \times \frac{W}{2r} \times 2C}$.

\subsection{SeA-block}

SeA-block is adopted for the segmentation-diagnosis feature interaction. The architecture of SeA-block is shown in Fig. \ref{fig:overall} (c). The proposed SeA-block contains an encoder and a decoder. The encoder embeds the diagnosis feature according to its affinity with segmentation feature, which is implemented with the multi-head dot-product attention mechanism (MHA) \cite{vaswani2017attention}. Formally, consider encoding a diagnosis feature $f_{d}$ with segmentation feature $f_{m}$, we use $f_{m}$ as \textit{query} and $f_{d}$ as \textit{key} and \textit{value} of the attention, which can be formulated as:
\begin{equation}\label{equation:fd}
     {\rm MHA} (q, k, v) = {\rm MHA} (f_{m} + E_{m},f_{d} + E_{d}, f_{d}),
\end{equation}
where $E_{m}, E_{d}$ are positional encodings \cite{carion2020end} for segmentation feature and diagnosis feature respectively. The features are all reshaped into a sequence of flattened patches following ViT. In this attention mechanism, the normalized affinity weights is first calculated between \textit{query} and \textit{key} to reflect the correlation between diagnosis and segmentation feature globally. Then the affinity weights are used to select and reinforce the diagnosis feature through the dot production of \textit{value}. After the attention, the Layer Normalization \cite{ba2016layer} with residual connection is applied before and after the MLP layer. The embedded diagnosis feature, which we denoted as $f_{md}$, is outputted with the same shape as the inputs. 

A decoder is connected after the encoder to map $f_{md}$ back to diagnosis feature space. There are two inputs for the decoder, diagnosis embedding $f_{md}$ and original diagnosis feature $f_{d}$. Being symmetrical to the encoder, decoder is implemented by the multi-head attention with diagnosis feature $f_{d}$ as $query$ and diagnosis embedding $f_{md}$ as $key$ and $value$, which can be formulated as:
\begin{equation}\label{equation:fd}
     {\rm MHA} (q, k, v) = {\rm MHA} (f_{d} + E_{d},f_{md} + E_{md}, f_{md}),
\end{equation}
where $E_{d}, E_{md}$ are positional encodings for diagnosis feature and diagnosis embedding respectively. The decoder transfers $f_{md}$ to a diagnosis feature by enhancing its affinity with $f_{d}$. A self-attention block is connected after the decoder to refine the representations. The obtained sequence will be reshaped back as a vitalized diagnosis feature $\hat{f}_{d} $ with the same shape as $f_{d}$.

\section{Experiment}
\subsection{Diagnosis Tasks}
We evaluate SeATrans on three different disease diagnosis tasks: glaucoma diagnosis, thyroid cancer diagnosis and melanoma diagnosis. Glaucoma is predicted from fundus images and is assisted by OD/OC segmentation. Thyroid cancer is predicted from ultrasound images and is assisted by the thyroid nodule segmentation. Melanoma is predicted from dermoscopic images and is assisted by skin lesions segmentation. The experiments of glaucoma, thyroid cancer and melanoma diagnosis are conducted on REFUGE-2 dataset \cite{fang2022REFUGE2}, TNMIX dataset  \cite{tn3k,tnscui} and ISIC dataset \cite{gutman2016skin}, which contain 1200, 8046, 1600 samples, respectively. The datasets are publicly available with both segmentation and diagnosis labels. Train/validation/test sets are split following the default settings of the dataset.

\subsection{Experimental Settings}
In our experiments, the main framework utilizes the UNet \cite{ronneberger2015u} architecture as the segmentation model and ResNet50 \cite{he2016deep} as the diagnosis model. The segmentation network is pre-trained on heterologous data distribution. All the experiments are implemented with the PyTorch platform and trained/tested on 4 Tesla P40 GPU with 24GB of memory. All images are uniformly resized to the dimension of 256$\times$256 pixels. The networks are trained in an end-to-end manner using Adam optimizer with a mini-batch of 16 for 80 epochs. The learning rate is initially set to 1 $\times 10^{-4}$. The detailed configurations can be found in the code.

To verify the effectiveness of SeATrans, we compare it with several baselines. The vanilla baseline is a standard classification model implemented by ResNet50 with no segmentation mask provided. Three other baselines are implemented by commonly used segmentation-assisted diagnosis techniques \cite{anwar2018medical}, which are denoted as 'Base-cat', 'Base-multi', and 'Base-ROI', respectively. 'Base-cat' concatenates the estimated masks with the raw images as the input of the diagnosis model. 'Base-multi' learns a single network for both segmentation and diagnosis. 'Base-ROI' crops the region of interest (ROI) based on the estimated segmentation masks. 

In order to verify the generalization of the models, we train segmentation network on homologous (-homo) and heterologous (-hetero) data, respectively. '-homo' means segmentation and diagnosis network are trained on the same source of data. '-hetero' means segmentation model is trained on an external dataset, which is RIGA \cite{almazroa2017agreement}, DDIT \cite{ddti} and PH2 \cite{mendoncca2013ph} for glaucoma, thyroid cancer and melanoma diagnosis, respectively.


\subsection{Main Results}
Comparing SeATrans with baselines in Table \ref{tab:baseline}, we can see significant improvement on all three diagnosis tasks. Concretely,
comparing with the best baseline by AUC, SeATrans improves 6.56\%, 6.78\% and 8.14\% on glaucoma, thyroid cancer and melanoma diagnosis respectively, indicating SeATrans can gain general and considerable improvement comparing with the present commonly used techniques. SeATrans also achieves the highest sensitivity with competitive accuracy and specificity, indicating it is more applicable to the real clinical scenarios, since sensitivity is commonly of great concern in clinical scenes. 

Comparing vanilla baseline with the other methods, we can see except 'Base-multi', the segmentation more or less improves the diagnosis performance. It demonstrates the segmentation information of lesions/tissues is definitely useful for the automated diagnosis models. However, the improvement it can bring depends largely on the way we use it. Multi-task learning based methods seemed to be invalid according to our experimental results. This may be due to the large discrepancy between segmentation and diagnosis features. The segmentation encoder extract the low-level structure features while the diagnosis needs the high-level semantic features, it is thus hard to learn the universal features in one encoder. SeATrans fuses the multi-scale segmentation features to first few layers of the diagnosis model. In this way, these structure-focused layers are enhanced by the awareness of lesions/tissues structures, and the later layers can still abstract the high-level diagnosis feature. As a result, SeATrans outperforms the other segmentation-assisted diagnosis methods by a large margin.

\begin{table}[b]
\centering
\caption{Comparing with the baselines. Accuracy, specificity, sensitivity and AUC (\%) are measured on three different diagnosis tasks.
Segmentation model performance measured by Dice score (\%) is also reported.}
\resizebox{\columnwidth}{!}{
\begin{tabular}{cc|cc|ccca|c|ccca|c|ccca}
\hline
\multicolumn{2}{c|}{Tasks}                                     & \multicolumn{6}{c|}{Glaucoma}                 & \multicolumn{5}{c|}{Thyroid Cancer}   & \multicolumn{5}{c}{Melanoma}          \\ \hline
\multicolumn{2}{c|}{Metrics}                                   & $\mathcal{D}_{disc}$    & $\mathcal{D}_{cup}$     & ACC   & SPE   & SEN   & AUC   & Dice  & ACC   & SPE   & SEN   & AUC   & Dice  & ACC   & SPE   & SEN   & AUC   \\ \hline
\multicolumn{1}{c|}{No Mask}                  & Baseline       & - & - & 82.95 & 94.06 & 37.97 & 77.29 & - & 79.29 & 93.75 & 68.62 & 77.08 & - & 78.53 & 92.23 & 22.22 & 72.49 \\ \hline
\multicolumn{1}{c|}{\multirow{4}{*}{-homo}}   & Base-cat   & 94.73 & 81.77 & 78.95 & 82.18 & 65.82 & 81.91 & 86.76 & 81.74 & \textbf{95.42} & 70.76 & 80.06 & 82.35 & 77.72 & 87.84 & 36.11 & 76.42 \\
\multicolumn{1}{c|}{}                         & Base-multi & 94.73 & 81.77 & 82.20 & \textbf{92.50} & 40.50 & 74.73 & 86.76 & 77.25 & 84.58 & 65.75 & 72.37 & 82.35 & 80.98 & \textbf{95.27} & 22.22 & 71.72 \\
\multicolumn{1}{c|}{}                         & Base-ROI   & 94.73 & 81.77 & 75.18 & 77.18 & 67.08 & 79.88 & 86.76 & 83.25 & 90.26 & 74.92 & 77.50 & 82.35 & 79.35 & 90.88 & 31.94 & 72.70 \\
\multicolumn{1}{c|}{}                         & SeATrans       & 94.73 & 81.77 & \textbf{86.96} & 90.93 & \textbf{70.88} & \textbf{88.47} & 86.76 & \textbf{85.54} & 91.75 & \textbf{78.84} & \textbf{86.84} & 82.35 & \textbf{85.56} & 93.77 & \textbf{62.74} & \textbf{84.56} \\ \hline
\multicolumn{1}{c|}{\multirow{4}{*}{-hetero}} & Base-cat   & 94.60 & 78.31 & 82.95 & 92.19 & 50.63 & 80.70 & 85.17 & 82.13 & 88.93 & 73.58 & 80.15 & 80.07 & 79.89 & 92.91 & 26.39 & 74.41 \\
\multicolumn{1}{c|}{}                         & Base-multi & 94.60 & 78.31 & 76.19 & 90.31 & 55.70 & 72.40 & 85.17 & 81.35 & \textbf{94.81} & 67.79 & 72.21 & 80.07 & 80.98 & \textbf{95.27} & 22.22 & 68.70 \\
\multicolumn{1}{c|}{}                         & Base-ROI   & 94.60 & 78.31 & \textbf{83.20} & \textbf{94.06} & 39.24 & 77.52 & 85.17 & 82.39 & 87.74 & 78.65 & 76.73 & 80.07 & 72.28 & 77.03 & 52.78 & 72.22 \\
\multicolumn{1}{c|}{}                         & SeATrans  & 94.60 & 78.31 & 80.20 & 80.62 & \textbf{78.48} & \textbf{87.61} & 85.17 & \textbf{84.60} & 90.26 & \textbf{80.45} & \textbf{86.23} & 80.07 & \textbf{84.42} & 87.20 & \textbf{57.35} & \textbf{83.16} \\ \hline
\end{tabular}}\label{tab:baseline}
\end{table}

To verify the generalization of the methods, we also conduct the experiment on heterologous data, where the segmentation model is pre-trained on external dataset. Due to the domain shift, the segmentation masks/features would be inferior to '-homo', thus disturb the diagnosis models. Comparing '-homo' with '-hetero', we can see a drop on the AUC performance over all of the methods. But SeATrans shows very competitive generalization ability, dropping only about 1\% AUC on '-hetero'. 



\subsection{Comparing with SOTA}
To demonstrate the advantage of SeATrans, we compare it with SOTA methods for segmentation-assisted diagnosis. 
Table \ref{tab:sota} quantitatively compare SeATrans with nine SOTA segmentation-assisted diagnosis methods.  

\textbf{SeATrans vs Transformers}. 
Present SOTA transformer-based diagnosis architectures: ConViT \cite{d2021convit} and Swin Transformer \cite{liu2021swin} are involved for the comparison. Segmentation masks are concatenated as the inputs of the models. It shows SeATrans clearly outperforms these transformer architectures, increases about 5.60\%, 5.82\% and 7.10\% AUC on glaucoma, thyroid and melanoma, respectively. It demonstrates a large proportion of the improvement comes from the proposed feature fusion strategy, but not the transformer-like architecture.

\textbf{SeATrans vs ROI}. We compare SeATrans with ROI based segmentation-assisted diagnosis methods: DualStage \cite{bajwa2019two} and DENet \cite{fu2018disc}. 
It shows \cite{bajwa2019two} only gains marginal improvement compared with vanilla baseline. Although \cite{fu2018disc} achieves better performance, it is only applicable on glaucoma diagnosis. SeATrans outperforms ROI based methods by an average 4\% AUC on a range of tasks.

\textbf{SeATrans vs Channel Attention}. We also compare SeATrans with SOTA channel attention based segmentation-assisted diagnosis methods: AGCNN  \cite{li2019attention} and ColNet \cite{zhou2019collaborative}, who adopted channel-attention to enhance the diagnosis feature by the segmentation masks/features. We observe that SeATrans can surpass AGCNN and ColNet by 6.31\% and 3.10\% AUC on glaucoma, 3.99\% and 2.40\% on thyroid cancer,and 4.39\% and 3.84\% on melanoma diagnosis, indicating the superiority of SeATrans comparing with regional-correlated channel attention.

\textbf{SeATrans vs Multi-task}. Multi-task learning methods MagNet  \cite{gupta2021mag} and CMSNET \cite{zhou2021multi} are involved for the comparison. SeATrans consistently outperforms both methods, especially on thyroid cancer diagnosis, which outperforms MagNet and CMSVNET by 11.16\% and 10.13\% AUC respectively.

\textbf{SeATrans vs Transfer-learning}. L2T-KT \cite{wu2020leveraging} uniquely processed the task by teacher-student based transfer learning and achieved competitive performance. Comparing the AUC, SeATrans outperforms L2T-KT by 2.23\%, 2.55\% and 2.66\% on glaucoma, thyroid cancer and melanoma diagnosis,  respectively. SeATrans also achieves better sensitivity-speficity trade-off than L2T-KT. For example, SeATrans achieves 79.66\% F1 score which surpasses 77.43\% F1 score of L2T-KT on glaucoma diagnosis.

\textbf{Heterologous data Generalization}. Comparing with '-homo' and '-hetero', we can see ROI-based methods (Dual-stage, DENet) show the best generalization, since they used less segmentation information than the others. SeATrans and Transformer-based methods (ConViT, Swin) also show competitive generalization capability, which drop only about 1\% AUC on a range of tasks. Channel-attention based methods (AGCNN, ColNet) are more sensitive since their regional correlated assumption is vulnerable to the domain shift. Thanks to the dynamic and global nature of SeATrans, it gains high performance with very competitive generalization ability comparing with the other methods.

\begin{table}[h]
\centering
\caption{Comparing with SOTA segmentation-assisted diagnosis methods. Accuracy, specificity, sensitivity and AUC (\%) are measured on three different diagnosis tasks.}
\resizebox{0.9\columnwidth}{!}{
\begin{tabular}{c|c|ccc a |ccc a |ccc a }
\hline
                          &           & \multicolumn{4}{c|}{Glaucoma}  & \multicolumn{4}{c|}{Thyroid Cancer}                & \multicolumn{4}{c}{Melanoma}  \\ \hline
                          &           & ACC   & SPE   & SEN   & AUC   & ACC                       & SPE   & SEN   & AUC   & ACC   & SPE   & SEN   & AUC   \\ \hline
\multirow{10}{*}{-homo}   & ConViT \cite{d2021convit} & 80.45 & 86.56 & 55.69 & 82.87 & 80.85                     & 90.31 & 64.67 & 81.02 & 79.89 & 90.87 & 34.72 & 77.46 \\
                          & Swin \cite{liu2021swin}      & 81.95 & 91.56 & 43.03 & 82.32 & 82.76                     & 85.70 & 73.82 & 80.34 & 80.76 & 89.11 & 47.29 & 76.75 \\
                          & DualStage \cite{bajwa2019two} & 80.20 & 90.31 & 39.24 & 80.37 & 79.56                     & 85.64 & 70.93 & 77.15 & 78.26 & 87.84 & 38.89 & 72.34 \\
                          & DENet \cite{fu2018disc}     & 80.04 & 85.00 & 59.49 & 84.70 & -                         & -     & -     & -     & -     & -     & -     & -     \\
                          & AGCNN \cite{li2019attention}     & 81.20 & 89.68 & 41.77 & 82.16 & 84.78                     & 88.69 & 71.05 & 82.85 & 82.60 & 92.20 & 43.83 & 80.17 \\
                          & ColNet \cite{zhou2019collaborative}    & 79.69 & 79.69 & 79.74 & 85.36 & \textbf{87.60}                     & \textbf{94.08} & 72.47 & 84.43 & 85.21 & \textbf{98.31} & 30.98 & 80.72 \\
                          & MagNet \cite{gupta2021mag}    & 83.20 & \textbf{94.06} & 39.24 & 77.52 & 78.91                     & 86.71 & 69.25 & 75.68 & 75.54 & 81.08 & 52.77 & 71.77 \\
                          & CMSNET \cite{zhou2021multi}    & 64.16 & 73.41 & 61.85 & 80.86 & 74.52                     & 87.03 & 67.32 & 76.71 & 82.88 & 98.32 & 17.14 & 78.55 \\
                          & L2T-KT \cite{wu2020leveraging}    & 80.20 & 80.62 & 78.48 & 86.24 & 81.49                     & 90.31 & 75.59 & 84.29 & 79.34 & 84.69 & 58.10 & 81.90 \\
                          & SeATrans  & \textbf{86.96} & 90.93 & \textbf{70.88} & \textbf{88.47} & \multicolumn{1}{l}{85.54} & 91.75 & \textbf{78.84} & \textbf{86.84} & \textbf{85.59} & 93.77 & \textbf{62.74} & \textbf{84.56} \\ \hline
\multirow{10}{*}{-hetero} & ConViT \cite{d2021convit} & 80.45 & 91.56 & 35.44 & 82.37 & 77.46                     & 93.75 & 60.11 & 81.02 & 79.34 & 85.42 & 54.79 & 76.55 \\
                          & Swin \cite{liu2021swin}      & 72.18 & 83.54 & 69.37 & 81.85 & 80.34                     & 83.45 & 76.18 & 80.34 & 79.89 & 87.11 & 50.68 & 75.28 \\
                          & DualStage \cite{bajwa2019two} & 75.18 & 67.08 & 77.18 & 80.22 & 80.70                     & 88.69 & 75.71 & 77.15 & 81.79 & 92.56 & 37.50 & 72.63 \\
                          & DENet \cite{fu2018disc}     & 78.94 & 84.81 & 77.50 & 84.12 & -                         & -     & -     & -     & -     & -     & -     & -     \\
                          & AGCNN \cite{li2019attention}     & 66.16 & 62.81 & 79.74 & 80.94 & 82.29                     & 93.41 & 70.58 & 82.85 & 76.08 & 78.98 & 64.38 & 77.38 \\
                          & ColNet \cite{zhou2019collaborative}    & 61.40 & \textbf{97.46} & 52.50 & 82.78 & \textbf{84.93}                     & 91.19 & 73.82 & 84.43 & 82.33 & \textbf{93.91} & 34.72 & 77.95 \\
                          & MagNet \cite{gupta2021mag}    & 70.42 & 70.31 & 70.88 & 75.44 & 78.15                     & 87.74 & 77.36 & 75.68 & 79.34 & 90.50 & 33.33 & 69.67 \\
                          & CMSNET \cite{zhou2021multi}    & 60.15 & 77.21 & 55.93 & 78.17 & 81.75                     & 84.60 & 78.14 & 76.71 & 72.82 & 73.73 & 69.01 & 76.39 \\
                          & L2T-KT \cite{wu2020leveraging}    & 79.95 & 85.00 & 59.49 & 84.98 & 83.89                     & \textbf{94.04} & 76.18 & 84.29 & 82.06 & 88.51 & 55.56 & 80.75 \\
                          & SeATrans  & \textbf{80.70} & 80.62 & \textbf{78.48} & \textbf{87.61} & 84.60                     & 90.26 & \textbf{80.45} & \textbf{86.23} & \textbf{84.42} & 87.20 & \textbf{57.35} & \textbf{83.16}\\
                          \hline
\end{tabular}}\label{tab:sota}
\end{table}

\subsection{Ablation study }
Ablation studies are performed over each component of SeATrans, including multi-scale, asymmetric interaction and SeA-Block, as listed in Table \ref{tab:ablation}. The experiments are conducted on glaucoma diagnosis task. Feature concatenation is adopted to replace SeA-block. In Table \ref{tab:ablation}, as we sequentially adding the proposed modules on vanilla baseline, the model performance is gradually improved. First, by applying multi-scale segmentation-diagnosis integration, the AUC value is increased by 2\% on homologous data while only 0.6\% on heterologous data. This indicates that multi-scale integration can improve the diagnosis performance with limited generalization. Then, the asymmetric multi-scale interaction is applied to further focus the integration on the low-level features, which boosts the AUC by a 3.53 \% and a 3.42\% on '-homo' and '-hetero' respectively. Finally, SeA-Block is utilized for the segmentation-diagnosis interaction. It can be observed the diagnosis performance is remarkably improved, which gains 5.09\% and 6.34\% AUC improvement on '-homo' and '-hetero', respectively. It indicates SeA-block gains significant and general improvement by its dynamic and global interaction.
\begin{table}[h]
\centering
\caption{Ablation study on glaucoma diagnosis task. The diagnosis performance is measured by AUC (\%)}
\begin{tabular}{ccc|cc}
\hline
Multi-scale & Asymmetric & SeA-block &-homo & -hetero \\ \hline
            &            &          & 77.29    & 77.29      \\
\checkmark             &            &          & 79.85    & 77.84      \\
\checkmark             &   \checkmark          &          & 83.38    & 81.27      \\
\checkmark             &    \checkmark         &  \checkmark         & \textbf{88.47}    & \textbf{87.61}      \\ \hline
\end{tabular}\label{tab:ablation}
\end{table}
\section{Conclusion}
In this work, we proposed SeATrans to overcome the shortcomings of existing segmentation-assisted diagnosis models. In SeATrans, asymmetric multi-scale interaction is proposed to address the segmentation-diagnosis scale level discrepancy. Then SeA-block is constructed for the global and dynamic feature interaction between segmentation and diagnosis space. Extensive empirical experiments demonstrated the general and superior performance of the proposed SeATrans on
a range of medical image diagnosis tasks.
\bibliographystyle{splncs04}
\bibliography{egbib}
\clearpage

\section{Supplementary material}
\begin{figure*}[h]
\centering
\includegraphics[width=1\textwidth]{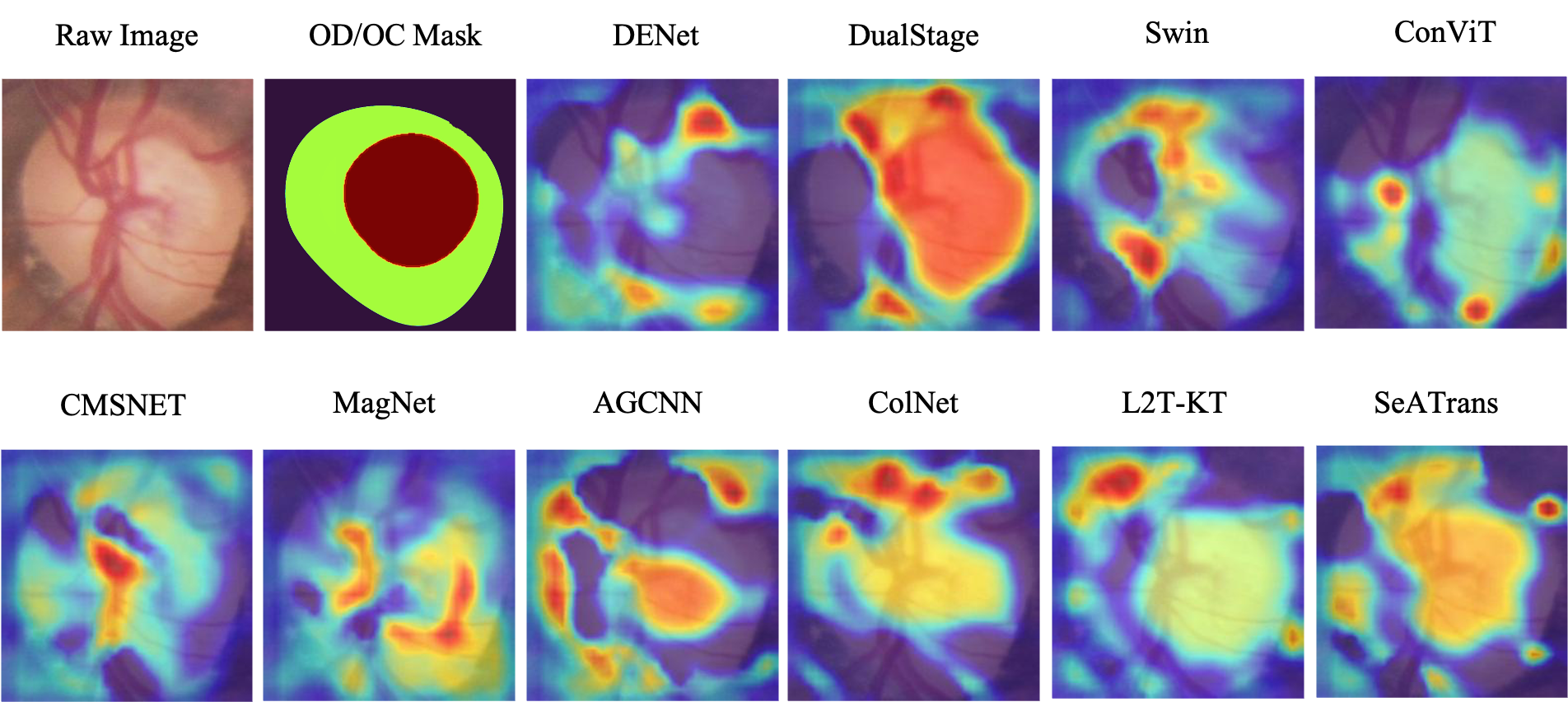}
\caption{Visualization results compared with the other methods on fundus images based glaucoma diagnosis. Grad-CAM is adopted to show the attentive regions for the diagnosis.}
\label{fig:seavis}
\end{figure*}

In order to further analysis the interrelation of the segmentation and the diagnosis. We adopt the network explanation techniques on the models to visualize the discriminative features. Grad-CAM\cite{selvaraju2017grad} is a commonly used explanation tool that produces visual explanations for model decisions. It visualize the gradients of the loss function with pixel-wise weighted feature maps. We compare the Grad-CAM produced visualization results on an glaucoma diagnosis example in Fig. \ref{fig:seavis}. 

We can see ROI based methods (DENet and DualStage) and Transformer based methods (Swin and ConViT) show less attention on the clinical focused region, like optic cup.
It may because these methods impose the segmentation enhancement on the model inputs rather than the deep features. Although the explanation is not so good, some of these models with sophisticated network structures still achieve fine diagnosis performance, like Swin, ConViT and DENet. Some of the recent literature also show that the sophisticated networks will show stronger capability but inferior explanation\cite{zhang2019theoretically,wu2019generating,ilyas2019adversarial}, since they are prone to learn some features that discriminative to the networks while meaningless to the human\cite{ilyas2019adversarial}. Multi-task based methods (MagNet and CMSNET) and channel attention based methods (AGCNN and ColNet) mainly focus on the optic-cup region, which is important for the clinical glaucoma diagnosis. But most of them are not implemented with sufficient learnable parameters, which cause they show inferior diagnosis performance. Transfer-learning based method (L2T-KT) and proposed SeATrans pay more attention on the optic-cup region. Besides optic-cup region, SeATrans also focuses on the gap between OC and OD boundary, which is another important parameter indicating glaucoma suspect clinically. Such visualization results demonstrate SeATrans can reach superior diagnosis performance with clear and reasonable explanation. 
\end{document}